\documentclass[conference]{IEEEtran}
\IEEEoverridecommandlockouts
\usepackage{cite}
\usepackage{amsmath,amssymb,amsfonts}
\usepackage{algorithmic}
\usepackage{graphicx}
\usepackage{textcomp}
\usepackage{xcolor}
\usepackage{hyperref} 
\usepackage{algorithm}

\usepackage{threeparttable}
\usepackage{makecell}
\usepackage{longtable}
\usepackage{multirow}
\usepackage[thicklines]{cancel}

\usepackage{rotating}
\usepackage{booktabs} % For formal tables
\usepackage[hang]{footmisc}

\def\BibTeX{{\rm B\kern-.05em{\sc i\kern-.025em b}\kern-.08em
    T\kern-.1667em\lower.7ex\hbox{E}\kern-.125emX}}
\begin{document}

\title{Adaptive Process-Guided Learning: An Application in Predicting Lake DO Concentrations
}
\author{
    \IEEEauthorblockN{
        Runlong Yu\textsuperscript{1}, 
        Chonghao Qiu\textsuperscript{1}, 
        Robert Ladwig\textsuperscript{2},
        Paul C. Hanson\textsuperscript{3},
        Yiqun Xie\textsuperscript{4}, 
        Yanhua Li\textsuperscript{5}, 
        Xiaowei Jia\textsuperscript{1}
    }
    \IEEEauthorblockA{
        \textsuperscript{1}\textit{University of Pittsburgh, }
        \textsuperscript{2}\textit{Aarhus University,} \textsuperscript{3}\textit{University of Wisconsin-Madison,} \\
        \textsuperscript{4}\textit{University of Maryland,} \textsuperscript{5}\textit{Worcester Polytechnic Institute}
    }
   \IEEEauthorblockA{
        \{ruy59,chq29,xiaowei\}@pitt.edu,
        rladwig@ecos.au.dk, pchanson@wisc.edu,  
        xie@umd.edu, yli15@wpi.edu
    }
}

\maketitle

\begin{abstract}
	
	This paper introduces a \textit{Process-Guided Learning (Pril)} framework that integrates physical models with recurrent neural networks (RNNs)  to enhance the prediction of dissolved oxygen (DO) concentrations in lakes, which is crucial for sustaining water quality and ecosystem health. Unlike traditional RNNs, which may deliver high accuracy but often lack physical consistency and broad applicability, the \textit{Pril} method incorporates differential DO equations for each lake layer, modeling it as a first-order linear solution using a forward Euler scheme with a daily timestep. 
	However, this method is sensitive to numerical instabilities. When drastic fluctuations occur, the numerical integration is neither mass-conservative nor stable. Especially during stratified conditions, exogenous fluxes into each layer cause significant within-day changes in DO concentrations. 
	To address this challenge, we further propose an \textit{Adaptive Process-Guided Learning (April)} model, which dynamically adjusts timesteps from daily to sub-daily intervals with the aim of mitigating the discrepancies caused by variations in entrainment fluxes. \textit{April} uses a generator-discriminator architecture to identify days with significant DO fluctuations and employs a multi-step Euler scheme with sub-daily timesteps to effectively manage these variations.
	We have tested our methods on a wide range of lakes in the Midwestern USA, and demonstrated robust capability in predicting DO concentrations even with limited training data. While primarily focused on aquatic ecosystems, this approach is broadly applicable to diverse scientific and engineering disciplines that utilize process-based models, such as power engineering, climate science, and biomedicine.

\end{abstract}


\section{Introduction}

The concentration of dissolved oxygen (DO) in lakes, as the indicator of water quality and ecosystem health, plays a key role in sustaining aquatic biodiversity and ensuring water safety for human consumption~\cite{wilson2010water}. DO concentrations are shaped not just by the exchange of oxygen between air and water, but also by the metabolic processes of primary production and respiration~\cite{yu2024evolution}. In deeper lakes, for instance, light scarcity and decreased mixing with the oxygen-rich surface can lower oxygen~\cite{solomon2013ecosystem,phillips2020time}. Temperature fluctuations impact oxygen solubility and biochemical activities~\cite{staehr2010lake}. Land use changes reshape DO patterns and metabolism phenology~\cite{jenny2016urban,woolway2021phenological}.
As articulated by Edward A. Birge one century ago~\cite{birge1906gases}: The fluctuations in a lake's oxygen illustrate its ``life cycle'' more clearly than many other ecological indicators. This is particularly evident in nutrient-rich eutrophic lakes, where algal blooms can significantly deplete oxygen, creating detrimental ``dead zones'' for aquatic life. 

Recognizing the importance of predicting DO concentrations, scientists in fields such as limnology, hydrology, meteorology, and environmental engineering have developed process-based models to simulate the dynamics of freshwater ecosystems. These models, which are designed to assess the impacts of both external and internal factors, often integrate hydrodynamic and water quality models~\cite{janssen2015exploring,ladwig2022long}. 
These models adhere to first-order principles like mass and energy conservation but often require numerous parameterizations or approximations due to limited physical understanding or complexity involved in modeling intricate processes, leading to inherent biases. Typically, the predictions from these models depend on qualitative parameterizations that use approximations based on factors such as morphometric and geographic information, weather conditions, trophic state, and watershed land use. As a result, these models frequently deliver sub-optimal predictive performance. Moreover, calibrating these process-based models tends to be highly time-consuming due to the complex interplay among parameters~\cite{beven2006manifesto}.

With advances in data collection driven by improved sensor technologies, there is a significant opportunity to systematically enhance the modeling of physical processes in these areas through machine learning (ML) methods. However, direct application of black-box ML models to scientific problems, such as predicting DO concentrations, often leads to serious false discoveries~\cite{karpatne2024knowledge,lazer2014parable,willard2022integrating} due to several  major challenges: 
\textit{1.~Data requirements}: Advanced ML models that can effectively represent spatial and temporal processes inherent in physical systems often outperform traditional empirical models (e.g., regression-based models) used by the science community. However, these advanced models require extensive training data, which is often scarce in practical settings.
\textit{2.~Physical consistency and generalizability}: Black-box ML models primarily identify statistical relationships between inputs and target variables solely from training data. With the absence of physics and limited coverage of training data distributions, the patterns extracted by ML models may significantly violate some established physical relationships (e.g., mass and energy conservation). 
Consequently, these models struggle to generalize to conditions not seen during training. For instance, an ML model trained with current climate data might not accurately predict conditions under future, warmer climate scenarios.

To address these challenges, we introduce a \textit{\textbf{Pr}ocess-Gu\textbf{i}ded \textbf{L}earning (\textbf{Pril})} framework, which aims to integrate physical knowledge encoded in physical models with the ML model to enhance the prediction of DO concentrations in lakes.  
The central idea of \textit{Pril} is to generalize the loss function of the ML model to include the differential equation that governs the DO dynamics for each lake layer. 
Specifically, the general ordinary differential DO equation describes the DO dynamics of each water layer (upper surface layer or lower bottom layer) in response to exogenous fluxes and entrainment fluxes based on the mass balance relationship. 
\textit{Pril} approximates each layer's ordinary differential DO equation as a discrete first-order linear forward differencing solution using an explicit forward Euler scheme with a daily timestep. When the lake water column is under mixed conditions, the model calculates the total DO concentration as a function of direct exogenous exchange.

Incorporating the ordinary differential DO equation as a physical loss function can help reduce the hypothesis space of the ML model to be consistent with the known physical relationship (i.e., mass conservation) and potentially improve its generalizability. Moreover, \textit{Pril} only requires predicted DO values to compute the physical loss and thus can implement it over data samples without true DO observations. 
However, a major concern of this method is its sensitivity to numerical instabilities. 
In particular, existing numerical approximations of differential equations (e.g., the Euler method, finite difference method~\cite{thomas2013numerical}) or the automatic differentiation approach~\cite{paszke2019pytorch,raissi2019physics} become less accurate when the observed target variable has drastic changes over the discrete timesteps. 
As a result, the numerical integration for simulating target DO dynamics can be neither mass-conservative nor stable when drastic DO fluctuations occur. 
For example, deep lakes in the Midwestern USA, often have stratified depth layers in summer, and it could happen that the volume of the hypolimnion (the lower bottom layer) can shrink rapidly to less than 10\% of their previous day's size, and DO mixes from the hypolimnion into the epilimnion (the upper surface layer). The Euler scheme, operating on a daily timestep, inaccurately assumes that DO concentrations in the hypolimnion remain constant throughout the day. In reality, exogenous fluxes can deplete the DO in the hypolimnion within half a day, leaving none for later transport. 
Similar numerical instability issues can also be observed when DO mixes from the epilimnion into the hypolimnion.

% In such cases, a flexible timestep is needed to avoid large discrepancies caused by variations in entrainment fluxes. 

To address the numerical instability issue, we further propose an \textit{\textbf{A}daptive \textbf{Pr}ocess-Gu\textbf{i}ded \textbf{L}earning (\textbf{April})} model, which improves stability and predictive performance by dynamically adjusting timesteps under different scenarios.  In the DO prediction problem, we aim to automatically divide daily to sub-daily intervals when drastic fluctuations occur to mitigate large discrepancies caused by variations in entrainment fluxes (causing volume changes for hypolimnion and epilimnion). As directly reducing the time step interval to sub-daily for the entire study period would significantly increase the computational cost, we introduce a novel generator-discriminator architecture to automatically identify the days with significant DO fluctuations and employ a multi-step Euler scheme with sub-daily timesteps to effectively manage these variations. Our contributions are summarized as follows:

\begin{itemize}
\item We address the prediction of DO concentrations in lakes by introducing a \textit{Pril} model that enriches the ML framework by incorporating the DO mass conservation.
\item We propose \textit{April}, which improves stability and performance by dynamically adjusting timesteps from daily to sub-daily intervals. This effectively mitigates numerical instabilities, ensuring physical consistency.
\item We tested our methods on a wide range of lakes in the Midwestern USA, demonstrating robust capability in predicting DO concentrations even with limited training data, and showing sensitivity to subtle variations.
\end{itemize} 
Our code is available at \href{https://github.com/RunlongYu/April}{https://github.com/RunlongYu/April}.

\begin{figure}[t]
\centerline{\includegraphics[width=\linewidth]{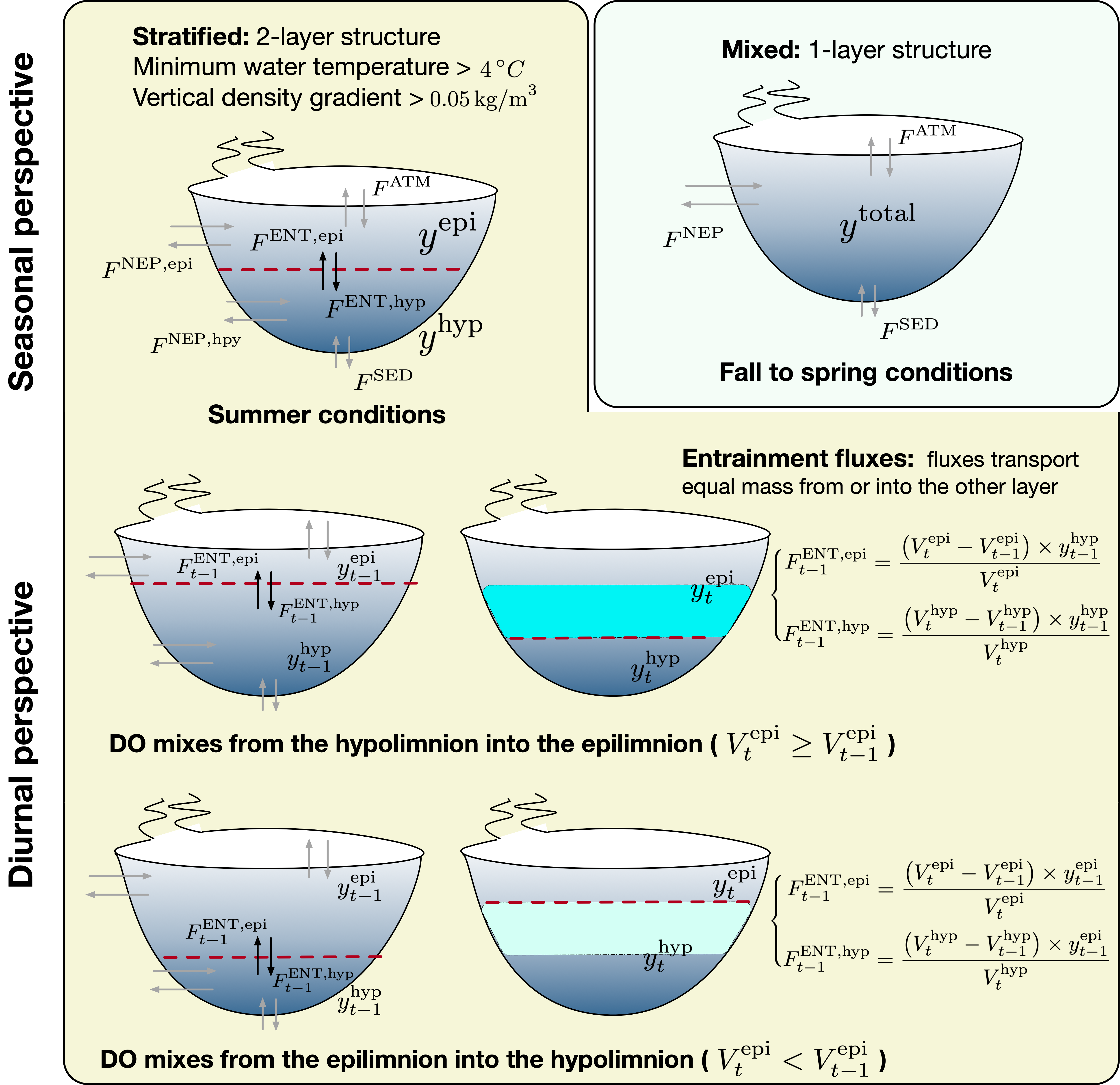}}
\caption{Seasonal variations and diurnal fluctuations in DO concentrations.}
% \vspace{-0.2cm}
\label{fig1}
\end{figure}

\section{Problem Formulation}

Our goal is to predict the DO concentration on a daily scale. As shown in Fig.~\ref{fig1}, during summer conditions, our study focuses on changes in ecosystem-scale metabolic fluxes in stratified lakes—characterized by a vertical density difference of over $0.05 \, \text{kg/m}^3$ between the surface and bottom layers, an average water temperature above $4\, ^\circ C$, and the presence of a thermocline. For analytical simplicity, we divide the water column into two distinct layers during the summer: the epilimnion (upper surface layer) and the hypolimnion (lower bottom layer), each with its own oxygen and metabolic kinetics. We treat the DO prediction for the epilimnion and hypolimnion as two tasks. From fall to spring, we treat the water column as completely mixed. 
In such mixed conditions, we aim to predict the total DO concentration in the lake.

For each lake, we have access to its phenological features $\pmb{x}_t$  on each date $t$. 
These features, spanning $m$ diverse fields $\pmb{x}_t = \{ x_t^1, \cdots, x_t^m \}$, encompass morphometric and geographic details; flux-related data; weather factors; a range of trophic states; and diverse land use proportions. 
In addition to these input features, we also have observed DO concentrations $y_t$ on certain days. During summer, these observations include separate DO concentrations for the epilimnion $y_{t}^{\rm epi}$ and hypolimnion $y_{t}^{\rm hyp}$. From fall to spring, these observations are the total DO concentration $y_{t}^{\rm total}$.

% \section{Preliminaries}

\section{Preliminaries on  Process-based Model} \label{Section3A}

The physical process-based model, as detailed in~\cite{ladwig2022long}, simulates the dynamics of DO concentration modeling various physical processes. 
These processes include the fluxes caused by atmospheric exchange ($F^{\rm ATM}$), net ecosystem production ($F^{\rm NEP}$), mineralization through sediment oxygen demand ($F^{\rm SED}$). Additionally, during the summer, it accounts for DO entrainment fluxes from or into the other layer driven by turbulent flow ($F^{\rm ENT}$), as depicted in Fig.~\ref{fig1}.
We refer to the fluxes caused by atmospheric exchange, net ecosystem production, and mineralization through sediment oxygen demand, among other factors, as exogenous flux. Turbulent forces drive entrainment fluxes that either shallow or deepen the thermocline, affecting the transport of DO into either the epilimnion or the hypolimnion.

The process-based DO model has several parameters, such as the rate of net ecosystem production and the idealized areal flux rate, which are often specifically calibrated for individual lakes when training data are available. The basic calibration method involves running the model for combinations of parameter values and selecting the parameter set that minimizes the error between the simulated and observed DO values. This calibration process can be both labor- and computationally intensive, and if executed without expert knowledge of parameter meanings and acceptable values, it can create model formulations that perform poorly when evaluated against test data. Furthermore, even in the ideal case of complete and error-free observational data, calibration processes are limited by the simplifications and rigid formulations of parameters in these physical process-based models.

% \subsection{Long-Short Term Memory (LSTM) Networks} \label{Section3B}

% The LSTM model~\cite{hochreiter1997long} establishes a transition relationship for the hidden representation $\pmb{h}_t$ via an LSTM cell. This cell integrates input features $\pmb{x}_t$ at each timestep with information carried forward from previous timesteps.

% To perform regression for continuous values, we generate the predicted DO concentration $\hat{y}_t$ at each timestep $t$ using a linear combination of hidden units:
% \begin{equation}
% 	\hat{y}_t = \pmb{W} \cdot \pmb{h}_t + \pmb{b},
% \end{equation} where $\pmb{W} $ and $\pmb{b}$ denote the weight and bias parameters, respectively.

% We apply the LSTM model to each task separately. For every instance, we incorporate a feature to denote whether conditions are mixed or stratified (0 for mixed, 1 for stratified) and another to distinguish between the epilimnion and hypolimnion layers (0 for the epilimnion, 1 for the hypolimnion). Given the true observation $y_t$ at every timestep, our training loss is defined as follows:
% \begin{equation}
% 	\mathcal{L}_{\rm RNN} = \frac{1}{|B|}\sum_{t \in B} \big(y_{t} -\hat{y}_{t} \big) ^2, 
% \end{equation} where $B$ denotes the set of instance indices in a mini-batch, and $\hat{y}_{t}$ denotes the predicted result given by the learned model. 

\section{Process-Guided Learning}

In this section, we introduce the \textit{Pril} framework, which integrates the DO mass flow into the ML training process. 
The idea is to define a physical loss based on the DO mass conservation and incorporate it into the training objective of the ML model. 
In contrast to the standard ML training loss $\mathcal{L}_{\rm ML}$ that measures the difference between observed $y_t$ and predicted $\hat{y}_t$, the physical loss $\mathcal{L}_{\rm MC}$ measures the discrepancy between the predicted $\hat{y}_t$ and the simulated target variable $\tilde{y}_t$ based on the mass balance process. Combining $\mathcal{L}_{\rm MC}$ and $\mathcal{L}_{\rm ML}$, our goal is to create predictions that match available sparse observations while penalizing those that significantly violate the mass conservation relationship. 
The new loss function is expressed as follows: 
% \begin{equation} 
% 	\label{eq1a}
% 	\mathcal{L}_{\textit{Pril}} = \mathcal{L}_{\rm ML} + \lambda_{\rm MC} \mathcal{L}_{\rm MC},
% \end{equation} 
% \begin{equation} 
% 	\label{eq1b}
% 	\mathcal{L}_{\rm ML} = \frac{1}{|B|}\sum_{t \in B} \big(y_{t} -\hat{y}_{t} \big) ^2,
% \end{equation} 
% \begin{equation} 
% 	\label{eq1c}
% 	\mathcal{L}_{\rm MC} = \frac{1}{T}\sum_{t=1}^T {\rm ReLU}\, \Big( | \hat{y}_{t} - \tilde{y}_t | -\tau_{\rm MC} \Big) ,  
% \end{equation} 
% \begin{equation} 
% 	\mathcal{L}_{\rm RNN} = \frac{1}{|B|}\sum_{t \in B} \big(y_{t} -\hat{y}_{t} \big) ^2, 
% \end{equation} 
\begin{equation} \small
	\label{eq9}
	\begin{aligned}
	\mathcal{L}_{\textit{\footnotesize Pril}} &= \mathcal{L}_{\rm ML} + \lambda_{\rm MC} \mathcal{L}_{\rm MC} , \\
 \mathcal{L}_{\rm ML} & = \frac{1}{|B|}\sum_{t \in B} \big(y_{t} -\hat{y}_{t} \big) ^2, \\
	\mathcal{L}_{\rm MC} &= \frac{1}{T}\sum_{t=1}^T {\rm ReLU}\, \Big( | \hat{y}_{t} - \tilde{y}_t | -\tau_{\rm MC} \Big) ,  
\end{aligned}
\end{equation} 
where $B$ denotes the set of instance with observed $y_t$, and $T$ denotes the total length of time series.  
We set a tolerance threshold $\tau_{\rm MC}$ for the mass conservation loss, which accommodates potential inaccuracies due to minor, unmodeled physical processes or observational errors in DO data. 
The function ${\rm ReLU(\cdot)}$ ensures that only discrepancies exceeding this threshold contribute to the conservation penalty.  
The hyper-parameter $\lambda_{\rm MC}$ controls the balance between the standard supervised loss and the mass conservation loss.

The incorporation of the physical loss $\mathcal{L}_{\rm MC}$ provides several benefits.
It reduces the search space of the ML model by regularizing the model to be consistent with known physical relationships, which can potentially enhance the model performance in data-sparse and out-of-sample scenarios. Moreover, the computation of  $\mathcal{L}_{\rm MC}$ does not require observed $y$ values and thus can be implemented on large unlabeled data points.

Note that \textit{Pril} is agnostic of specific ML-based models. In this work, we choose Long-Short Term Memory (LSTM) networks as the base model. The choice is justified by its superior performance in previous hydrology studies~\cite{jia2019physics,jia2021simlr,hanson2020predicting,chen2023physics}, in which it is shown to outperform advanced ML models due to its effectiveness in capturing temporal dependencies, especially given sparse real observations. We also test other advanced models (e.g., Transformer) in the experiments (Section~\ref{sec:exp}).

In the DO prediction problem, $\tilde{y}_t$ is simulated by combining the predicted DO concentration from the previous day and the DO fluxes.
Fig.~\ref{fig1} illustrates the major fluxes that govern the diurnal fluctuations of DO concentration in lakes. The interactions among these fluxes result in changes to the DO mass within the lake. Our method approximates the general ordinary differential equation for DO using a discrete, first-order linear forward differencing solution:
\begin{equation} 
\frac{d {y}_t}{dt} = \sum_{i}^{n} F^i_{t},
\end{equation} where $F^i_{t}$ denotes the $i^{\rm th}$ flux among $n$ total fluxes at time $t$, which can either increase or decrease the DO concentration. In the following, we will describe the simulation of $\tilde{y}_t$ under two different scenarios, (i) the well-mixed condition from fall to spring, and (ii) the stratified condition during summer.

\textit{\textbf{Well-mixed condition:}} Assuming that diurnal variations in the total lake volume are negligible when the lake water column is under mixed conditions, the dynamics can be modeled using a forward Euler scheme with a daily timestep $\Delta t = 1$, given as:
\begin{equation} 
	\tilde{y}_{t}^{\rm total} = \hat{y}_{t-1}^{\rm total} + F^{\rm EXO}_{t-1} \times \Delta t,  
	\label{eq4}
\end{equation}
where $F^{\rm EXO}$ represents the exogenous fluxes, including $F^{\rm ATM}$, $F^{\rm NEP}$, and $F^{\rm SED}$ (depicted as gray arrows in Fig.~\ref{fig1}). 

% where $\Delta t$ is the daily timestep. %, set to $1$ for daily timestep.  

\begin{figure*}[!t]
	\centerline{\includegraphics[width=0.87\linewidth]{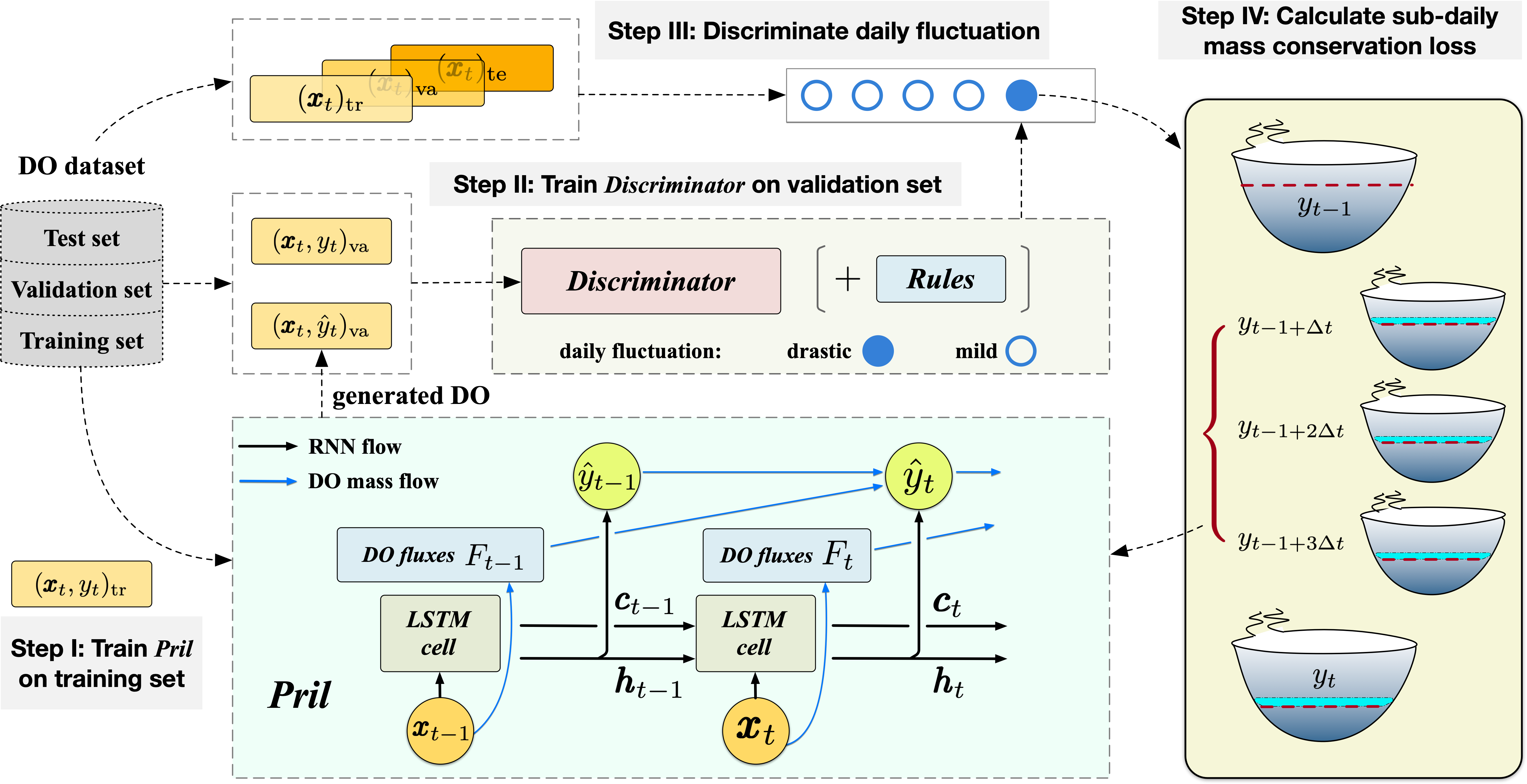}}
	\caption{Overall framework of the \textit{April} model.}
	\vspace{-0.4cm}
	\label{fig2}
\end{figure*}

\textit{\textbf{Stratified condition:}} When addressing the stratified conditions prevalent during summer, the dynamics become considerably more complex. It is necessary to consider not only the daily volume changes in the epilimnion and hypolimnion layers but also the entrainment fluxes from or into the other layer driven by turbulent flow $F^{\rm ENT}$. Consequently, we adjust Eq. (\ref{eq4}) accordingly:
\begin{equation}
	\tilde{y}_{t}^{\rm epi} = \left( \hat{y}_{t-1}^{\rm epi} + F^{\rm EXO,epi}_{t-1} \times \Delta t\right) \times \frac{V_{t-1}^{\rm epi}}{V_t^{\rm epi}} + F^{\rm ENT,epi}_{t-1},
	\label{eq5}
\end{equation} 
\begin{equation}
	\tilde{y}_{t}^{\rm hyp} = \left( \hat{y}_{t-1}^{\rm hyp} + F^{\rm EXO,hyp}_{t-1} \times \Delta t\right) \times \frac{V_{t-1}^{\rm hyp}}{V_t^{\rm hyp}} + F^{\rm ENT,hyp}_{t-1},
	\label{eq6}
\end{equation} 
where $V^{\rm epi}_t$ and $V^{\rm hyp}_t$ represent the volume of epilimnion and hypolimnion, respectively. 

There are two scenarios for changes in entrainment flux. When DO mixes from the hypolimnion into the epilimnion (i.e., $V^{\rm epi}_t \ge V^{\rm epi}_{t-1}$), the entrainment fluxes are given as:
\begin{equation} \small
	\left\{
	\begin{aligned}
		F^{\rm ENT,epi}_{t-1} &= \frac{\big(V_t^{\rm epi}-V_{t-1}^{\rm epi}\big) \times \hat{y}_{t-1}^{\rm hyp}}{V_t^{\rm epi}} \\
		F^{\rm ENT,hyp}_{t-1} &= \frac{\big(V_t^{\rm hyp}-V_{t-1}^{\rm hyp}\big) \times \hat{y}_{t-1}^{\rm hyp}}{V_t^{\rm hyp}}
	\end{aligned}
	\right. .
	\label{eq7}
\end{equation} Conversely, when DO mixes from the epilimnion into the hypolimnion (i.e., $V^{\rm epi}_t < V^{\rm epi}_{t-1}$), the calculations adjust to:
\begin{equation} \small
	\left\{
	\begin{aligned}
		F^{\rm ENT,epi}_{t-1} &= \frac{\big(V_t^{\rm epi}-V_{t-1}^{\rm epi}\big) \times \hat{y}_{t-1}^{\rm epi}}{V_t^{\rm epi}} \\
		F^{\rm ENT,hyp}_{t-1} &= \frac{\big(V_t^{\rm hyp}-V_{t-1}^{\rm hyp}\big) \times \hat{y}_{t-1}^{\rm epi}}{V_t^{\rm hyp}}
	\end{aligned}
	\right. .
	\label{eq8}
\end{equation} Note that $\big(V_t^{\rm epi}-V_{t-1}^{\rm epi}\big) = -\big(V_t^{\rm hyp}-V_{t-1}^{\rm hyp}\big)$, indicating that both fluxes transport equal mass from or into their respective opposite layers.

\section{Adaptive Process-Guided Learning}

%\subsection{Limitations of the \textit{Pril} Model}

The \textit{Pril} model incorporates the differential DO equation for each lake layer, modeling it as a first-order linear solution using a forward Euler scheme with a daily timestep. However, this type of numerical method is sensitive to numerical instabilities, particularly during significant fluctuations observed in stratified conditions. We will first analyze why numerical instabilities could occur in two common scenarios. Then we propose a generator-discriminator method to address this issue through adaptive timestep adjustment. Finally, we will discuss refining the \textit{Pril} method based on the adaptive adjustment.

\subsection{Analysis of numerical instabilities} \label{subsec:5a}

\textit{\textbf{Scenario A:}}
When the volume of the hypolimnion rapidly decreases (i.e., $V_{t-1}^{\rm hyp} \gg V_{t}^{\rm hyp}$), DO mixes from the hypolimnion to the epilimnion. The Euler scheme with daily timestep may inaccurately assume constant DO concentrations in the hypolimnion, $\hat{y}_{t-1}^{\rm hyp}$, throughout the day. Combining Eqs.~(\ref{eq6}) and (\ref{eq7}), we can simulate $\tilde{y}_{t}^{\rm hyp}$ as follows:
\begin{equation} 
	\tilde{y}_{t}^{\rm hyp} = \hat{y}_{t-1}^{\rm hyp} + F^{\rm EXO,hyp}_{t-1} \times \Delta t  \times \frac{V_{t-1}^{\rm hyp}}{V_t^{\rm hyp}}.
	\label{eq10}
\end{equation} 
In reality, DO concentration is a dynamic variable that changes continuously, i.e.,  $\tilde{y}_{t-1+\Delta t}^{\rm hyp} (\Delta t < 1)$ can deviate from $\hat{y}_{t-1}^{\rm hyp}$ due to persistent exogenous and entrainment fluxes over the sub-daily intervals $\Delta t$. 
Often, the hypolimnion consumes oxygen faster than it is replenished, leading to frequent negative values for $F^{\rm EXO,hyp}_{t-1}$. Consequently, with $\frac{V_{t-1}^{\rm hyp}}{V_{t}^{\rm hyp}}$ being large, the second term in Eq.~(\ref{eq10}) becomes profoundly negative. This rapid depletion of DO in the hypolimnion can occur within half a day, leaving minimal oxygen for subsequent transport. If we assume the DO concentration in the hypolimnion remains constant throughout the day and use Eq.~(\ref{eq7}) to calculate $F^{\rm ENT,epi}_{t-1}$, then %incorporating this into Eq.~(\ref{eq5}) can result in a significant overestimation of 
the simulation %of $\hat{y}_{t}^{\rm epi}$ 
by Eq.~(\ref{eq5}) could significantly overestimate $\tilde{y}_{t}^{\rm epi}$. Given the DO mass conservation of the entire lake, $\tilde{y}_{t}^{\rm hyp}$ becomes underestimated.

\textit{\textbf{Scenario B:}} When there is a sharp decrease in the volume of the epilimnion (i.e., $V_{t-1}^{\rm epi} \gg V_{t}^{\rm epi}$), DO mixes from the epilimnion into the hypolimnion. Assuming that the epilimnion DO concentrations, $\hat{y}_{t-1}^{\rm epi}$, remain constant throughout the day, we integrate Eqs.~(\ref{eq5}) and (\ref{eq8}) to calculate $\tilde{y}_{t}^{\rm epi}$:
\begin{equation} 
	\tilde{y}_{t}^{\rm epi} = \hat{y}_{t-1}^{\rm epi} + F^{\rm EXO,epi}_{t-1} \times \Delta t  \times \frac{V_{t-1}^{\rm epi}}{V_t^{\rm epi}}.
	\label{eq11}
\end{equation} 
Rapid changes in the epilimnion, whether in terms of oxygen replenishment or consumption, can lead to substantial fluctuations, resulting in either large positive or negative values for $F^{\rm EXO,epi}_{t-1}$. Given the considerable ratio $\frac{V_{t-1}^{\rm epi}}{V_{t}^{\rm epi}}$, this can cause the second term in Eq.~(\ref{eq11})  to become either extremely positive or negative. If it is positive, then maintaining %the assumption that the 
a constant epilimnion DO concentration %is constant 
throughout the day can lead to significant underestimate of $\tilde{y}_{t}^{\rm hyp}$ following Eqs.~(\ref{eq8}) and (\ref{eq6}). 
Given the DO mass conservation of the entire lake, $\tilde{y}_{t}^{\rm epi}$ becomes overestimated.

\subsection{Adaptive timestep adjustment}
The aforementioned two scenarios illustrate the need for a flexible timestep to effectively minimize the significant discrepancies caused by variations in entrainment fluxes.  In response, we introduce the \textit{April} model, as depicted in Fig.~\ref{fig2}.

The \textit{April} model consists of two main components: a DO generator $G$ and a discriminator $D$. The generator $G$ creates DO prediction  $\hat{y}_t$ given $\pmb{x}_t$  using the standard \textit{Pril} method. The generator is trained under the assumption that only mild fluctuations exist, which can be effectively managed with a forward Euler scheme using a daily timestep. The discriminator aims to distinguish between days with mild fluctuations ($t \in \mathbb{N}$) and those with drastic fluctuations ($t \in \mathbb{P}$). 
The training labels ($\mathbb{N}$ or $\mathbb{P}$) are created by referring to the difference between the generator output $G(\pmb{x}_t)$ and observation $y_t$. Specifically, the drastic fluctuations are identified as the days for which the 
the standard $\textit{Pril}$ with daily timestep cannot accurately resemble observed $y_t$, i.e., $\mathbb{P}=\{t\,| \,\,|G(\pmb{x}_t)-y_t|>\gamma\}$, where $\gamma$ is a threshold set as 1.5 times overall root mean squared error in our tests. 
In this implementation, \textit{April} employs a $\omega$-parameterized multi-layer network  $D_{\omega}(\pmb{x}_t)$, designed to optimize the following problem:
\begin{equation}
	\max_{\omega} \;\;  \mathbb{E}_{t \in \mathbb{N}} \left[ \log D_{\omega} \left( \pmb{x}_t \right) \right] 
	+  \mathbb{E}_{t \in \mathbb{P}} \left[ \log \left( 1- D_{\omega} \left( \pmb{x}_t \right) \right) \right]. 
	\label{eq12}
\end{equation} 
% Here, $D_{\omega}(\pmb{x}_t)$ evaluates the likelihood that the generated output $G(\pmb{x}_t)$ %$\hat{y}_t$, generated from the day’s input $\pmb{x}_t$, 
% accurately resembles $y_t$. Essentially, it performs a binary classification task using cross-entropy loss. \textcolor{red}{The training labels $\mathbb{E}$ and $\mathbb{D}$ are created on days with consecutive observations based on XXXX criteria...}

We train the generator on the training dataset until convergence, and then train the 
discriminator on the validation set. Once fully trained, the discriminator can assess whether the DO fluctuations for a given day are mild or drastic across the entire dataset. 
Additionally, to improve identification accuracy, we integrate specific rules into the classification, including the inclusion of days with daily volume changes exceeding 20\%, which significantly influence DO flux changes.

% Two considerations should be noted:
% \begin{enumerate}
% 	\item Employing sub-daily timesteps throughout would significantly increase computational demands due to the complex interactions among parameters.
% 	\item The procedure for modeling DO mass flow does not rely on true labels or observations. DO fluxes are calculated solely based on input drivers and the predicted DO concentrations.
% \end{enumerate}

\subsection{Refinement of mass conservation loss}

Employing sub-daily timesteps throughout the entire time series could improve the simulation accuracy but would significantly increase computational demands due to the complex interactions among parameters. 
Effective identification of numerical instabilities allows us to refine the mass conservation loss %calculate specific sub-daily mass conservation losses according to these classifications. 
%To refine the mass conservation loss calculations 
from daily to sub-daily intervals only for days with drastic fluctuations. In particular,  we divide each identified day into $k$ parts, establishing a sub-daily timestep size of $\Delta t = \frac{1}{k}$. By linearly interpolating the volume from $t-1$ to $t$, we determine the volume change per sub-daily timestep as:
\begin{equation} \small
	\Delta V^{\rm epi}_t = \frac{1}{k} (V^{\rm epi}_t - V^{\rm epi}_{t-1}),
	\label{eq13}
\end{equation}
\begin{equation} \small
	\Delta V^{\rm hyp}_t = \frac{1}{k} (V^{\rm hyp}_t - V^{\rm hyp}_{t-1}).
	\label{eq14}
\end{equation} Note that $\Delta V^{\rm epi}_t = -\Delta V^{\rm hyp}_t$, indicating that both fluxes transport equal mass from or into their respective opposite layers. For any sub-daily intervals $0 \le i < k$, the entrainment flux between the time intervals $t-1+i\Delta t$ and $t-1+(i+1)\Delta t$ can be determined by modifying Eqs.~(\ref{eq7}) and (\ref{eq8}). Specifically, when DO mixes from the hypolimnion into the epilimnion, the entrainment fluxes for each sub-daily timestep are given~as:
\begin{equation} \small
	\left\{
	\begin{aligned}
		F^{\rm ENT,epi}_{t-1+i\Delta t} &= \frac{\Delta V^{\rm epi}_t \times \tilde{y}_{t-1+i\Delta t}^{\rm hyp}}{V_{t-1+(i+1)\Delta t}^{\rm epi}} \\
		F^{\rm ENT,hyp}_{t-1+i\Delta t} &= \frac{\Delta V^{\rm hyp}_t \times \tilde{y}_{t-1+i\Delta t}^{\rm hyp}}{V_{t-1+(i+1)\Delta t}^{\rm hyp}}
	\end{aligned}
	\right. .
	\label{eq15}
\end{equation} Conversely, when DO mixes from the epilimnion into the hypolimnion, the corresponding entrainment fluxes for each sub-daily timestep are adjusted accordingly:
\begin{equation} \small
	\left\{
	\begin{aligned}
		F^{\rm ENT,epi}_{t-1+i\Delta t} &= \frac{\Delta V^{\rm epi}_t \times \tilde{y}_{t-1+i\Delta t}^{\rm epi}}{V_{t-1+(i+1)\Delta t}^{\rm epi}} \\
		F^{\rm ENT,hyp}_{t-1+i\Delta t} &= \frac{\Delta V^{\rm hyp}_t \times \tilde{y}_{t-1+i\Delta t}^{\rm epi}}{V_{t-1+(i+1)\Delta t}^{\rm hyp}}
	\end{aligned}
	\right. .
	\label{eq16}
\end{equation}

\begin{algorithm} [!t]
\small
	\caption{Multi-step Euler scheme with sub-daily timesteps}
	\label{alg1}
	\textbf{Input}: Predicted DO concentrations $\hat{y}_{t-1}^{\rm epi}$ and  $\hat{y}_{t-1}^{\rm hyp}$, exogenous fluxes, \mbox{epilimnion and hypolimnion volumes, sub-daily timestep size $\Delta t = \frac{1}{k}$}
 \vspace{-0.30cm}
\begin{algorithmic}
        \STATE $\big(\tilde{y}_{t-1}^{\rm epi}, \tilde{y}_{t-1}^{\rm hyp}  \big) \leftarrow \big(\hat{y}_{t-1}^{\rm epi}, \hat{y}_{t-1}^{\rm hyp} \big) $
        \vspace{0.12cm}
	\FOR{$i = 0$ to $k-1$}
	\IF{$V^{\rm epi}_t \ge V^{\rm epi}_{t-1}$}
	\STATE Calculate $F^{\rm ENT,epi}_{t-1+i\Delta t}$, $F^{\rm ENT,hyp}_{t-1+i\Delta t}$ via Eq.~(\ref{eq15}) 
	\ELSE
	\STATE Calculate $F^{\rm ENT,epi}_{t-1+i\Delta t}$, $F^{\rm ENT,hyp}_{t-1+i\Delta t}$ via Eq.~(\ref{eq16}) 
	\ENDIF
	\STATE $\tilde{m}^{\rm epi}  \leftarrow   \tilde{y}_{t-1+i\Delta t}^{\rm epi} \cdot V_{t-1+i\Delta t}^{\rm epi} + F^{\rm EXO,epi}_{t-1} \cdot \Delta t \cdot V_{t-1}^{\rm epi} $ 
	\vspace{0.12cm}
	\STATE $\tilde{y}_{t-1+(i+1)\Delta t}^{\rm epi} \leftarrow \frac{\tilde{m}^{\rm epi}}{ {V_{t-1+(i+1)\Delta t}^{\rm epi}}} + F^{\rm ENT,epi}_{t-1+i\Delta t}$
	\vspace{0.18cm}
	\STATE $\tilde{m}^{\rm hyp}  \leftarrow   \tilde{y}_{t-1+i\Delta t}^{\rm hyp} \cdot V_{t-1+i\Delta t}^{\rm hyp} + F^{\rm EXO,hyp}_{t-1} \cdot \Delta t \cdot V_{t-1}^{\rm hyp} $
	\vspace{0.12cm}
	\STATE $\tilde{y}_{t-1+(i+1)\Delta t}^{\rm hyp} \leftarrow \frac{\tilde{m}^{\rm hyp}}{ {V_{t-1+(i+1)\Delta t}^{\rm hyp}}} + F^{\rm ENT,hyp}_{t-1+i\Delta t}$
	\vspace{0.08cm}
	\ENDFOR
	\RETURN $\tilde{y}_{t}^{\rm epi}$, $\tilde{y}_{t}^{\rm hyp}$
\end{algorithmic}
\end{algorithm}

\newcolumntype{L}[1]{>{\raggedright\arraybackslash}p{#1}}
\newcolumntype{C}[1]{>{\centering\arraybackslash}p{#1}}
\newcolumntype{R}[1]{>{\raggedleft\arraybackslash}p{#1}}

\begin{table*}[t]\footnotesize
	\centering
	\caption{Comparative performance of DO concentration ($g / m^{3}$) prediction in terms of RMSE during summer conditions.}
        \vspace{-0.1cm}
	\begin{tabular}{L{1.5cm}C{1.55cm}C{1.55cm}C{1.55cm}C{1.55cm}C{1.55cm}C{1.55cm}C{1.55cm}C{1.55cm}}
		\toprule
		\multicolumn{1}{l}{\multirow{2}[3]{*}{Algo. Name}} & \multicolumn{2}{c}{Cluster - \textit{Kohlman Lake}} & \multicolumn{2}{c}{Cluster - \textit{Lake Minnetonka}} & \multicolumn{2}{c}{Cluster - \textit{Gervais Lake}} & \multicolumn{2}{c}{Cluster - \textit{Bde Maka Ska}} \\
		\cmidrule(lr){2-3} \cmidrule(lr){4-5} \cmidrule(lr){6-7} \cmidrule(lr){8-9}
		& Epi. & Hyp. & Epi. & Hyp. & Epi. & Hyp. & Epi. & Hyp. \\
		\midrule
		Process & 1.677 \textcolor{gray}{(0.000)} & 1.859 \textcolor{gray}{(0.000)} & 1.656 \textcolor{gray}{(0.000)} & 2.317 \textcolor{gray}{(0.000)} & 1.503 \textcolor{gray}{(0.000)} & 2.077 \textcolor{gray}{(0.000)} & 1.686 \textcolor{gray}{(0.000)} & 2.089 \textcolor{gray}{(0.000)} \\
		LSTM & 1.770 \textcolor{gray}{(0.015)} & 1.945 \textcolor{gray}{(0.042)} & 1.567 \textcolor{gray}{(0.020)} & 2.139 \textcolor{gray}{(0.045)} & 1.537 \textcolor{gray}{(0.034)} & 1.925 \textcolor{gray}{(0.041)} & 1.725 \textcolor{gray}{(0.054)} & 1.940 \textcolor{gray}{(0.012)} \\
		EA-LSTM & 1.776 \textcolor{gray}{(0.026)} & 1.838 \textcolor{gray}{(0.031)} & 1.650 \textcolor{gray}{(0.024)} & 2.118 \textcolor{gray}{(0.014)} & 1.513 \textcolor{gray}{(0.007)} & 1.947 \textcolor{gray}{(0.030)} & 1.619 \textcolor{gray}{(0.039)} & 1.886 \textcolor{gray}{(0.047)} \\
		Transformer & 1.749 \textcolor{gray}{(0.016)} & 1.824 \textcolor{gray}{(0.052)} & 1.592 \textcolor{gray}{(0.007)} & 2.162 \textcolor{gray}{(0.010)} & 1.502 \textcolor{gray}{(0.050)} & 1.923 \textcolor{gray}{(0.021)} & 1.705 \textcolor{gray}{(0.018)} & 1.963 \textcolor{gray}{(0.015)} \\
		TFT & 2.034 \textcolor{gray}{(0.131)} & 2.024 \textcolor{gray}{(0.023)} & 1.606 \textcolor{gray}{(0.019)} & 2.384 \textcolor{gray}{(0.059)} & 1.613 \textcolor{gray}{(0.031)} & 2.188 \textcolor{gray}{(0.085)} & 1.690 \textcolor{gray}{(0.019)} & 2.440 \textcolor{gray}{(0.147)} \\
		\midrule
		\textit{Pril} & 1.726 \textcolor{gray}{(0.044)} & 1.803 \textcolor{gray}{(0.029)} & 1.571 \textcolor{gray}{(0.021)} & 2.058 \textcolor{gray}{(0.045)} & 1.515 \textcolor{gray}{(0.033)} & 1.904 \textcolor{gray}{(0.062)} & 1.631 \textcolor{gray}{(0.183)} & 1.790 \textcolor{gray}{(0.047)} \\
		\textit{April} & \textbf{1.535} \textcolor{gray}{(0.031)} & \textbf{1.796} \textcolor{gray}{(0.023)} & \textbf{1.547} \textcolor{gray}{(0.026)} & \textbf{2.024} \textcolor{gray}{(0.012)} & \textbf{1.418} \textcolor{gray}{(0.013)} & \textbf{1.850} \textcolor{gray}{(0.015)} & \textbf{1.512} \textcolor{gray}{(0.008)} & \textbf{1.707} \textcolor{gray}{(0.007)} \\
		\bottomrule
	\end{tabular}%
	\vspace{-0.3cm}
	\label{table01}
\end{table*}

Using a multi-step Euler scheme with sub-daily timesteps, we model DO dynamics as shown in Algorithm~\ref{alg1}. In this model, $\tilde{m}^{\rm epi}$ and $\tilde{m}^{\rm hyp}$ serve as tentative variables that represent the DO mass for the epilimnion and hypolimnion, respectively.
The volumes $V_{t-1+i\Delta t}^{\rm epi}$, $V_{t-1+(i+1)\Delta t}^{\rm epi}$, $V_{t-1+i\Delta t}^{\rm hyp}$, and $V_{t-1+(i+1)\Delta t}^{\rm hyp}$ are calculated through linear interpolation.
The exogenous fluxes, $F^{\rm EXO,epi}_{t-1}$  and $F^{\rm EXO,hyp}_{t-1}$, are multiplied by the respective volumes, $V_{t-1}^{\rm epi}$ and $V_{t-1}^{\rm hyp}$, because these exogenous fluxes are generated by the simulation at time $t-1$ and use the volumes at $t-1$ to calculate density.

On days with drastic DO fluctuations, we replace Eq.~(\ref{eq5}) and Eq.~(\ref{eq6}) with the outputs from Algorithm~\ref{alg1}. Using this updated information, Eq.~(\ref{eq9}) allows us to calculate the revised DO mass conservation loss. \textit{April} effectively mitigates numerical instabilities,
ensuring adherence to the law of mass conservation for physical consistency.

\section{Experimental Evaluation} 
\label{sec:exp}

% We conduct extensive experiments across a variety of lakes in the Midwestern USA, to investigate the research questions: 
% \begin{itemize}
% 	\item \textbf{RQ1.} How does the effectiveness of the proposed \textit{Pril} and \textit{April} models compare to other baseline methods?
% 	\item \textbf{RQ2.} What is the performance of \textit{Pril} and \textit{April} in time-series analysis of DO concentrations?
% 	\item \textbf{RQ3.} To what extent do \textit{Pril} and \textit{April} ensure physically consistent solutions?
% 	\item \textbf{RQ4.} How sensitive are \textit{Pril} and \textit{April} to changes in their key hyperparameters?
% \end{itemize}

\begin{figure}[t]
	\centerline{\includegraphics[width=0.84\linewidth]{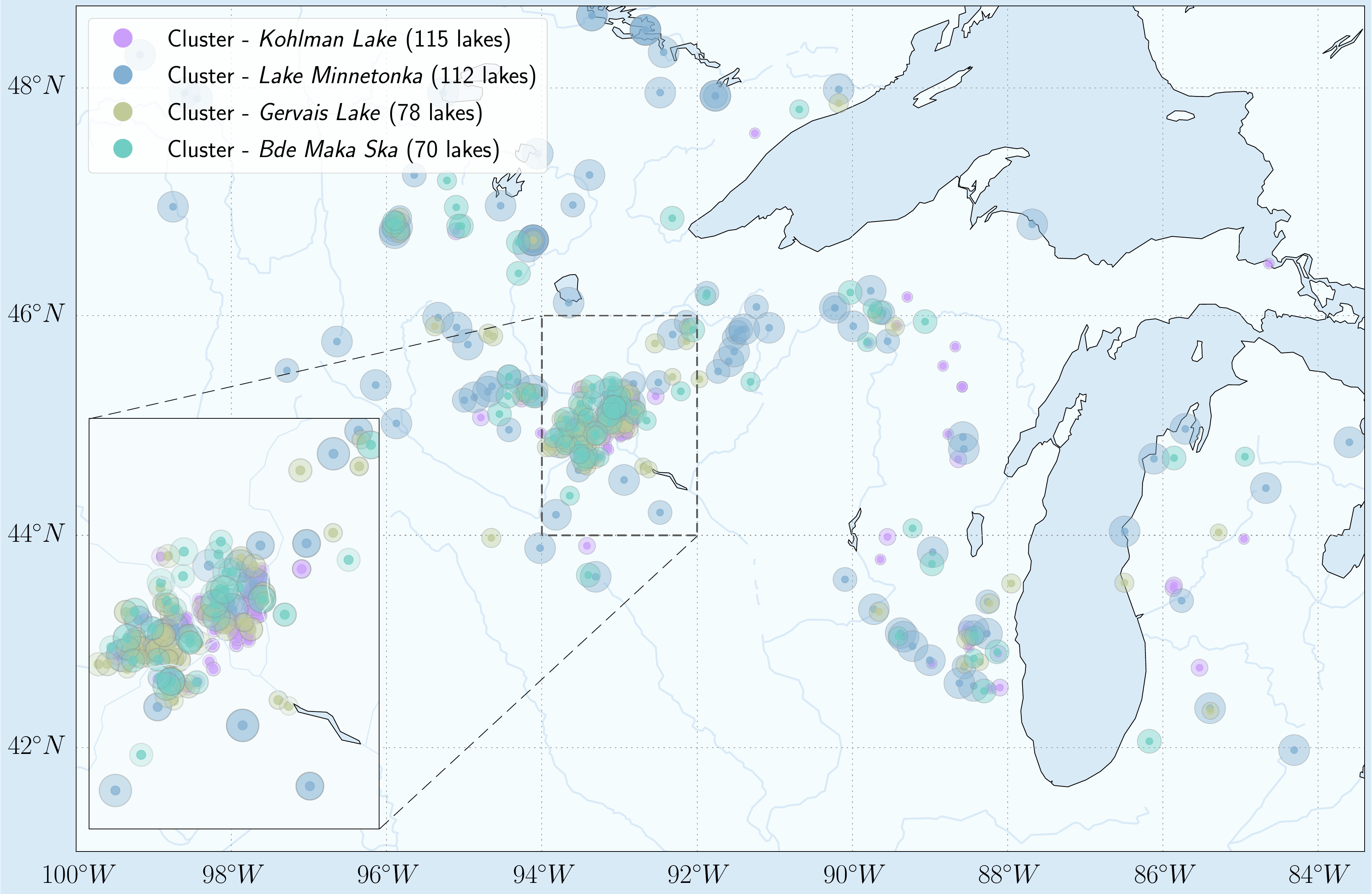}}
 \vspace{-0.16cm}
	\caption{Clustering of lakes based on depth and surface area.}
	\label{fig3}
\end{figure}

\subsection{Data preparation} 

We evaluate the proposed methods for predicting DO concentration using a comprehensive dataset. This dataset spans 41 years and includes ecological observations from 375 lakes across the Midwestern USA, beginning in 1979. It includes around 5.58 million daily records, each characterized by 39 fields of phenological features. These features encompass morphometric details, weather conditions, trophic states, and land use data. 
Observed DO data were sourced from the Water Quality Portal (WQP). Lake residence time was taken from the HydroLAKES dataset. Trophic state probabilities (eutrophic, oligotrophic, dystrophic) were from a dataset referenced in~\cite{meyer2024national}. Land use proportions of each lake's watershed were taken from the National Land Cover Database (NLCD). 
Of these, 37,986 records include observed DO concentrations for both the epilimnion and hypolimnion layers, primarily used for predicting summer conditions. Additionally, another 6,547 records contain observed DO concentrations relevant to mixed conditions, applicable from fall to spring.

\begin{table}[t]\footnotesize
        \vspace{-0.3cm}
	\setlength{\tabcolsep}{2.7pt} % Adjust the number to decrease/increase the space between columns
	\centering
	\caption{Comparative performance during fall to spring.}
        \vspace{-0.1cm}
	\begin{tabular}{L{1.50cm}C{1.54cm}C{1.54cm}C{1.54cm}C{1.54cm}}
		\toprule
		\multicolumn{1}{l}{\multirow{2}[3]{*}{Algo. Name}} & \multicolumn{1}{c}{Cluster - \textit{KL}} & \multicolumn{1}{c}{Cluster - \textit{LM}} & \multicolumn{1}{c}{Cluster - \textit{GL}} & \multicolumn{1}{c}{Cluster - \textit{BMS}} \\
		\cmidrule(lr){2-2} \cmidrule(lr){3-3} \cmidrule(lr){4-4} \cmidrule(lr){5-5}
		& Total & Total & Total & Total \\
		\midrule
		Process & 4.336 \textcolor{gray}{(0.000)} & 1.696 \textcolor{gray}{(0.000)} & 2.963 \textcolor{gray}{(0.000)} & 1.766 \textcolor{gray}{(0.000)}  \\
		LSTM & 2.923 \textcolor{gray}{(0.211)} & 1.833 \textcolor{gray}{(0.068)} & 2.559 \textcolor{gray}{(0.117)} & 1.911 \textcolor{gray}{(0.101)}  \\
		EA-LSTM & 3.147 \textcolor{gray}{(0.082)} & 1.923 \textcolor{gray}{(0.102)} & 2.746 \textcolor{gray}{(0.226)} & 1.780 \textcolor{gray}{(0.048)}  \\
		Transformer & 3.226 \textcolor{gray}{(0.187)} & 2.009 \textcolor{gray}{(0.039)} & 2.623 \textcolor{gray}{(0.114)} & 1.906 \textcolor{gray}{(0.053)} \\
		TFT & 5.531 \textcolor{gray}{(1.110)} & 2.071 \textcolor{gray}{(0.108)} & 3.598 \textcolor{gray}{(0.098)} & 1.940 \textcolor{gray}{(0.145)}  \\
		\midrule
		\textit{Pril} & 2.753 \textcolor{gray}{(0.246)} & 1.755 \textcolor{gray}{(0.059)} & \textbf{2.435} \textcolor{gray}{(0.071)} & \textbf{1.697} \textcolor{gray}{(0.030)}  \\
		\textit{April} & \textbf{2.278} \textcolor{gray}{(0.254)} & \textbf{1.552} \textcolor{gray}{(0.066)} & 2.523 \textcolor{gray}{(0.069)} & 1.718 \textcolor{gray}{(0.037)}  \\
		\bottomrule
	\end{tabular}%
	\vspace{0.2cm}
	\label{table02}
\end{table}

\begin{figure*}[!t]
	\centerline{\includegraphics[width=0.93\linewidth]{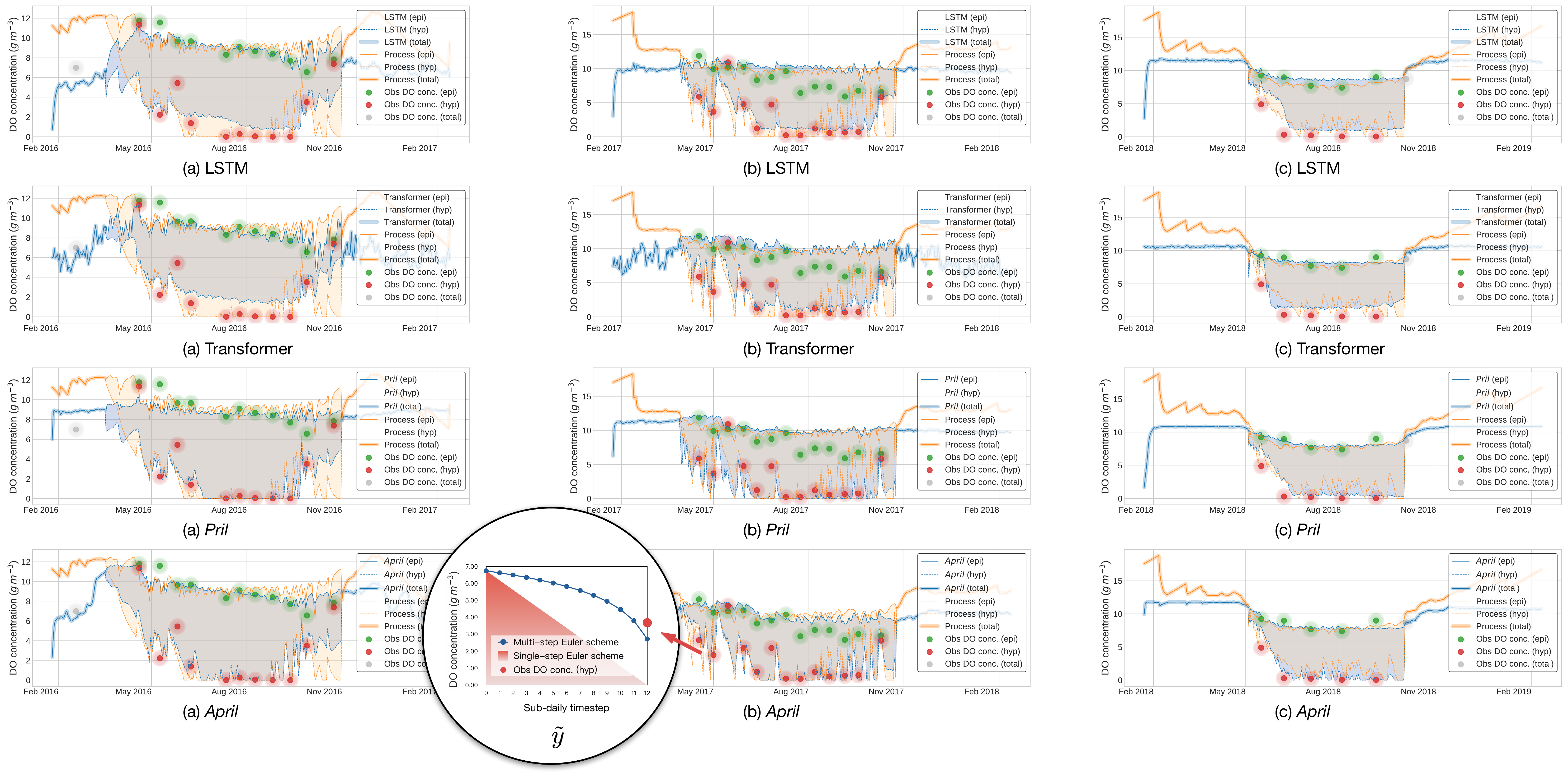}}
	\vspace{-0.25cm}
	\caption{Time-series analysis of DO concentrations: comparing predictions, physical process-based simulation, and observed values.}
	\vspace{-0.36cm}
	\label{fig4}
\end{figure*}

We classify lakes based on characteristics that significantly influence oxygen dynamics, such as surface area and depth. Larger surface areas facilitate more effective oxygen exchange with the atmosphere, while deeper lakes often exhibit increased oxygen consumption in deeper waters due to restricted surface replenishment. To systematically organize this diverse collection of lakes, we employ a balanced K-means clustering algorithm to form uniformly sized lake clusters~\cite{malinen2014balanced}, ensuring an equitable distribution across types. Lakes are categorized into four distinct groups based on depth and surface area, with clustering results depicted in Fig.~\ref{fig3}. The intensity of the point's color indicates the lake's depth, and the size of the point reflects the lake's surface area. Typically, lakes within the same category share similar depths and surface areas. Given the high concentration of lake observation points in  Minnesota, we have named each cluster after a prominent lake in Minnesota that typifies the group: Kohlman Lake, Lake Minnetonka, Gervais Lake, and Bde Maka Ska. For model development, the dataset is segmented as follows: data collected up to 2011 form the training set, data spanning 2012 to 2015 constitute the validation set, and data from 2016 to 2019 are designated as the test set.

\subsection{Baselines} 

In order to demonstrate the effectiveness of our models, we conduct comparisons against a set of baseline methods, namely Process, LSTM, EA-LSTM, Transformer, and TFT. Each is described as follows:
\begin{itemize}
	\item \textsl{Process}~\cite{ladwig2022long}: Discussed in Section~\ref{Section3A}, this baseline is the state-of-the-art physical process-based model. It serves as the guiding model for the machine learning approaches we propose. The flux features utilized in our models are derived by calibrating this model with observed data.
	\item \textsl{LSTM}~\cite{hochreiter1997long}: The LSTM model is a standard RNN configuration. It does not incorporate a mass conservation loss, distinguishing it from our methodological enhancements.
	\item \textsl{EA-LSTM}~\cite{kratzert2019towards}: As an Entity-Aware LSTM, this model integrates hydrological behavior and distinguishes between similar dynamic behaviors (such as wind speed and temperature), across various entities (including distinct morphometric characteristics). 
	\item \textsl{Transformer}~\cite{vaswani2017attention}: The Transformer uses self-attention mechanisms to process sequence data simultaneously, enhancing its ability to handle long-range dependencies efficiently. Renowned for its speed and versatility, it excels in natural language processing.
	\item \textsl{TFT}~\cite{lim2021temporal}: The Temporal Fusion Transformer features an advanced architecture that merges time-series processing with gating mechanisms and attention components. 
\end{itemize} 

\begin{figure*}[!t]
	\centerline{\includegraphics[width=0.87\linewidth]{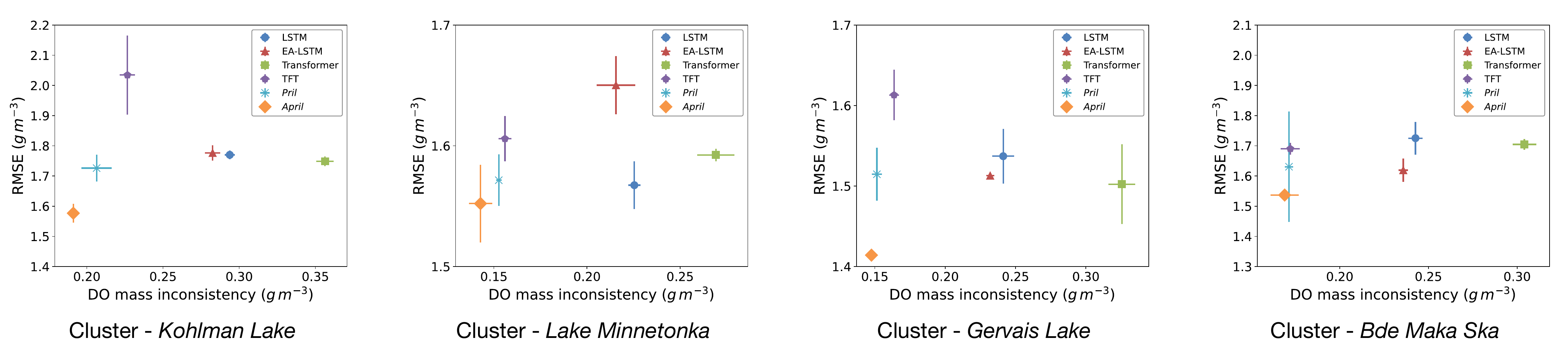}}
	\vspace{-0.25cm}
	\caption{Physical consistency analysis of epilimnion DO concentration predictions: evaluating RMSE and DO mass inconsistency.}
	\vspace{-0.35cm}
	\label{fig5}
\end{figure*}

\begin{figure*}[!t]
	\centerline{\includegraphics[width=0.87\linewidth]{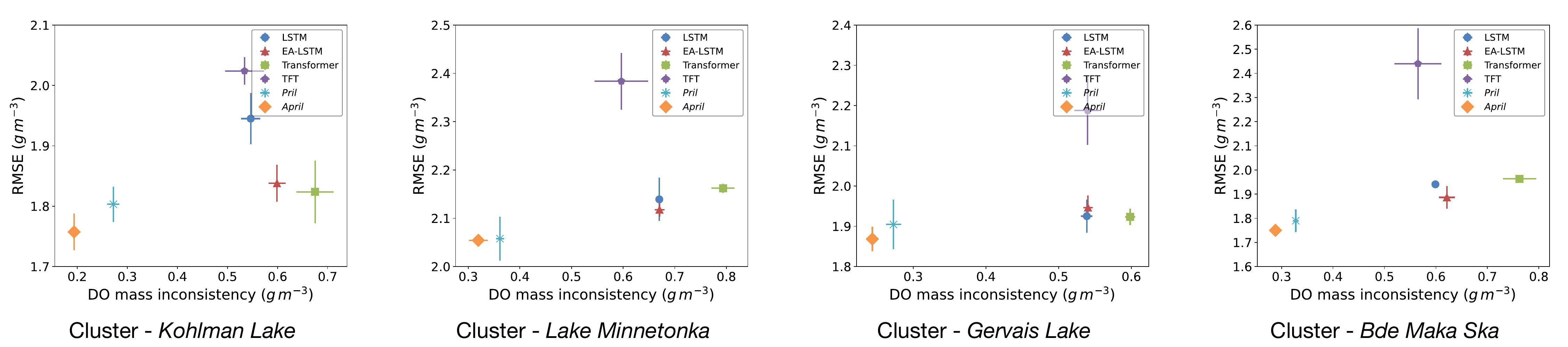}}
	\vspace{-0.25cm}
	\caption{Physical consistency analysis of hypolimnion DO concentration predictions: evaluating RMSE and DO mass inconsistency.}
	\vspace{-0.40cm}
	\label{fig6}
\end{figure*}

\vspace{-0.2cm}
\subsection{Implementation details}

For a fair comparison, all methods were implemented within the same coding framework. LSTM, EA-LSTM, Transformer, \textit{Pril}, and \textit{April} underwent initial pre-training on observed data from all lakes within each cluster before being fine-tuned on individual lakes. For TFT, due to limited data preventing effective convergence, we supplemented the training with simulated labels from the Process model on days lacking observed data, alongside actual observed data. 
Hyperparameters for all methods, including batch size (ranging from $8$ to $32$), learning rate (from $0.001$ to $0.05$), and hidden layer dimensions (from $20$ to $200$), were meticulously optimized through extensive grid searching, with the best configurations selected based on performance on validation data. For the Transformer and TFT, the number of attention heads was also varied (from $4$ to $16$). Specifically for \textit{Pril}, the mass conservation threshold $\tau_{\rm MC}$ was chosen from $\{0, 0.01, 0.05, 0.1, 0.5\}$, and the mass conservation weight $\lambda_{\rm MC}$ for both epilimnion and hypolimnion was chosen from $0$ to $1000$. For \textit{April}, the sub-daily timestep size $1/k$ was set to $1/12$, corresponding to two hours per step.

\vspace{-0.1cm}
\subsection{Experimental results} 

\subsubsection{Performance comparison}

Table~\ref{table01} provides a comparative analysis of \textit{Pril} and \textit{April} against baseline methods for predicting DO concentrations in the epilimnion and hypolimnion layers during summer stratified conditions. Table~\ref{table02} evaluates the performance of these models in predicting the total DO concentration under mixed conditions from fall to spring. Both tables measure performance based on the root mean square error (RMSE), including both mean and standard deviation (indicated in grey), calculated over five runs. From the results, we have the following key observations:
\begin{enumerate}
	\item Comparative analysis of Process, LSTM, and EA-LSTM reveals they show comparable performance across three tasks in four clusters. Process slightly underperforms, while EA-LSTM has a slight edge, illustrating the benefits of integrating hydrological behaviors into models.
	\item Comparing attention-based models (Transformer and TFT) with Process, LSTM, and EA-LSTM, Transformer performs comparably to EA-LSTM. Despite its success in NLP, the Transformer's advantage is modest in this context, limited by its point-to-point modeling which fails to capture the continuity inherent in time series data. The sparsity of DO data further impacts the effectiveness of models with extensive parameters, particularly affecting TFT, which ranks as the least effective among the baselines. 
	\item When comparing our proposed models, \textit{Pril} and \textit{April}, with other baselines, we observe significant improvements. Even without \textit{April}, \textit{Pril} outperforms all baselines in seven out of twelve cases, highlighting the substantial benefits of integrating physical consistency into the learning model. When including \textit{April}, our algorithms deliver the best performance across all three tasks in all four clusters. Notably, \textit{April} excels in predicting DO concentrations in both the epilimnion and hypolimnion layers during summer stratified conditions, underscoring the effectiveness of dynamically adjusting timesteps from daily to sub-daily intervals. \textit{April} and \textit{Pril} perform comparably in predicting total DO concentrations under mixed conditions from fall to spring. The multi-step Euler scheme with sub-daily timesteps significantly enhances model accuracy during summer when large fluctuations occur but has less impact from fall to spring.
\end{enumerate}

\begin{figure*}[!t]
	\centerline{\includegraphics[width=\linewidth]{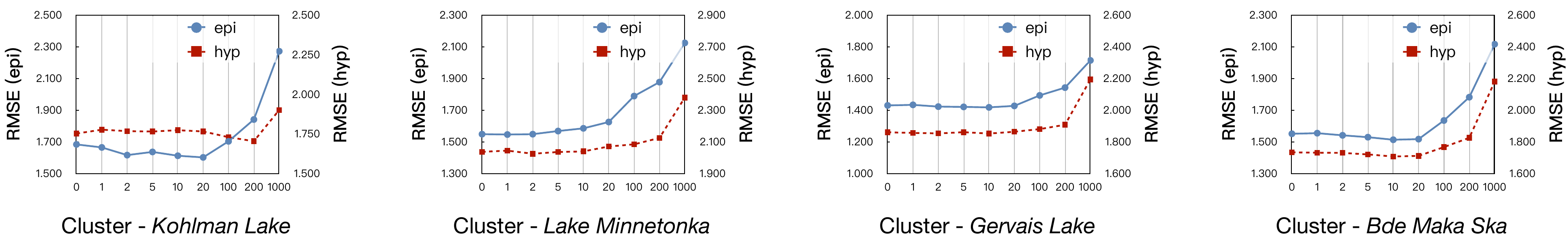}}
	\vspace{-0.25cm}
	\caption{Sensitivity analysis: Effects of adjusting epilimnion mass conservation weight $\lambda_{\rm MC}^{\rm epi}$ with fixed hypolimnion weight. } 
	 \vspace{-0.3cm}
	\label{fig7}
\end{figure*}

\begin{figure*}[!t]
	\centerline{\includegraphics[width=\linewidth]{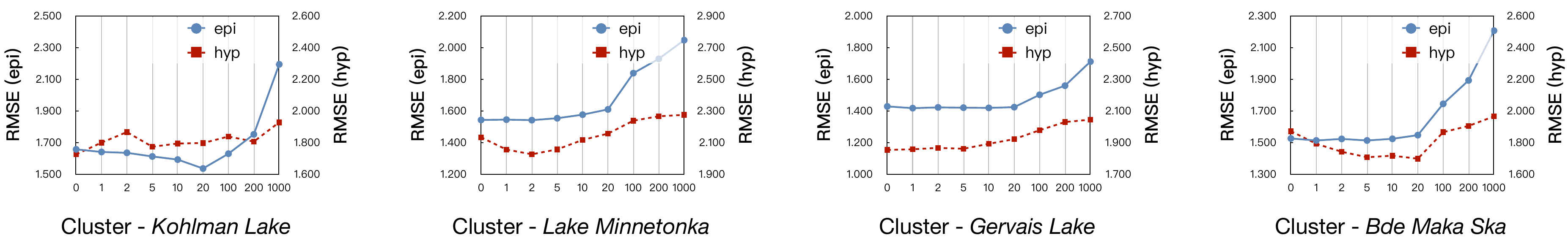}}
	\vspace{-0.25cm}
	\caption{Sensitivity analysis: effects of adjusting hypolimnion mass conservation weight $\lambda_{\rm MC}^{\rm hyp}$ with fixed epilimnion weight. }
	 \vspace{-0.45cm}
	\label{fig8}
\end{figure*}

\subsubsection{Time-series analysis} Fig.~\ref{fig4} offers a time-series comparison of DO predictions from LSTM, Transformer, \textit{Pril}, and \textit{April} against both physical process-based simulations and observed values. Given the limited availability of observed DO data from fall to spring and the heightened concern for DO concentration in lakes during summer (when the hypolimnion often experiences oxygen depletion, potentially leading to aquatic organism fatalities), this figure specifically focuses on results from the summer season of the testing period.

The analysis indicates that, in comparison to LSTM and Transformer, our models, \textit{Pril} and \textit{April}, not only closely align with observed values but also effectively detect subtle fluctuations. Their sensitivity is highlighted as both models accurately track the trends seen in physical process-based simulations. In contrast, LSTM and Transformer, though performing adequately in terms of RMSE, fail to capture critical dynamics such as the depletion of bottom oxygen in summer, rendering their results less practical. 

Upon detailed comparison of \textit{Pril} and \textit{April}, it becomes clear that \textit{April} more consistently aligns with observed values than \textit{Pril}. To demonstrate this, we have included a plot of $ \tilde{y}$ in Fig.~\ref{fig4}~\textbf{\textsf{(b) \emph{April}}}, which shows the DO mass conservation loss on a day with drastic fluctuations. Here, $\tilde{y}$ is calculated as the sum of the previous day's predicted DO concentration and the DO fluxes from the right-hand side of the Euler scheme equations (as described in Eq.~(\ref{eq9})). This demonstrates that when DO mixes from the hypolimnion to the epilimnion, the single-step Euler scheme tends to overestimate the epilimnion DO concentration and underestimate hypolimnion DO concentration compared to the multi-step Euler scheme. This observation is consistent with the analysis discussed in \textit{Scenario A} in Section~\ref{subsec:5a}.

\subsubsection{Physical consistency analysis} 

In scientific applications, machine learning models are expected to not only align with observed data but also maintain physical consistency. To demonstrate how \textit{Pril} and \textit{April} contribute to a physically consistent solution, Figs.~\ref{fig5} and~\ref{fig6} display the RMSE and DO mass inconsistency for each method's predictions of epilimnion and hypolimnion DO concentrations, respectively. The Process model is excluded from this inconsistency analysis since its DO mass inconsistency is consistently zero. 

We can observe that, across all clusters and tasks, \textit{Pril} and \textit{April} consistently position nearest to the bottom left corner, demonstrating their superior capability to reduce both prediction RMSE and DO mass inconsistency. A direct comparison between \textit{Pril} and \textit{April} shows that \textit{April} excels in both accuracy and physical consistency. Notably, the Transformer exhibits the highest DO mass inconsistency across all clusters and tasks. This underscores that reliance solely on data-driven machine learning may lead to significant misinterpretations in scientific applications.

\subsubsection{Sensitivity test} 

We conduct sensitivity tests to assess the impact of hyperparameters on \textit{April}, focusing on the mass conservation weight $\lambda_{\rm MC}$, a pivotal hyperparameter due to its direct role in ensuring alignment with physical laws. Figs.~\ref{fig7} and \ref{fig8} illustrate performance variations when one lake layer's weight is fixed while adjusting the opposite layer's weight.

Three key insights are derived from the analysis: First, utilizing excessively large values for $\lambda_{\rm MC}$ leads to increased RMSE, as it overshadows the standard supervised training loss.
Second, performance fluctuations are sharper for the epilimnion compared to the hypolimnion, amplified by the inherently higher DO concentrations in the epilimnion.
Third, adjustments to $\lambda_{\rm MC}$ in one layer consistently affect fluctuations in the opposite layer. This is due to the entrainment flux that mutually influences DO mass across both layers.

\section{Related Work} 

\textit{\textbf{Physics-based models.}}
The simulation of complex environmental processes has long been the domain of physics-based models, including different components in water cycles~\cite{hipsey2019general,markstrom2012p2s} and plant growth in agroecosystems~\cite{grant2010changes,zhou2021quantifying}. However, these models are necessarily approximations of reality and often rely on approximations and parameterizations~\cite{gupta2014debates,lall2014debates,mcdonnell2014debates}. 
With the recent advances in ML, there is a huge opportunity for modeling environmental ecosystems using the rapidly growing 
Earth observation data and ground observations about water, plants, soils, and climate. In particular, prior research has shown the promise of ML-based approaches in modeling agroecosystems~\cite{liu2022kgml,ghazaryan2020crop,van2020crop} and freshwater ecosystems~\cite{rahmani2021exploring,willard2021predicting}. 
Importantly, the outcomes of these models can be used to inform critical actions (e.g., distribution of subsidies \cite{national2018improving}) to mitigate natural disturbance-incurred food and freshwater shortages,  
which is necessary for continued sustainability and stability. 
Given the limitations of both physics-based and ML-based models, the improvement of prediction accuracy made through the proposed research would significantly benefit many societally relevant decision-making activities.

\textit{\textbf{DO concentration prediction.}}
Aquatic ecosystem models (AEMs) have been pivotal in the aquatic ecosystem science domain, helping us understand the complex interactions within ecosystems~\cite{janssen2015exploring}. 
These models blend hydrodynamics, water quality, and ecosystem processes, employing various methodologies such as GLM~\cite{hipsey2019general}, MyLake~\cite{saloranta2007mylake}, and LAKE2.0~\cite{stepanenko2016lake}. More advanced vertical one-dimensional AEMs including GLM-AED~\cite{hipsey2019general}, WET~\cite{nielsen2017open}, and PCLake~\cite{janssen2019pclake+} further enhance modeling capabilities.
Despite their comprehensive nature, AEMs encounter limitations due to their complexity, computational intensity, and equifinality issues, constraining their flexibility and wider applicability~\cite{luo2018autocalibration}. Simpler diel metabolism models, while insightful for short-term DO fluctuations, often overlook long-term ecosystem predictions due to missing hydrodynamic factors~\cite{staehr2012metabolism,giling2017delving,appling2018overcoming}. The integration of physics-based models with ML, particularly utilizing simulated data to enhance ML training where observed data are limited, holds transformative potential for bridging these gaps.

\section{Conclusion} 

In this paper, we proposed \textit{Pril} and \textit{April},  innovative machine learning models that integrate physical process consistency into scientific discovery.  \textit{Pril} incorporates DO mass conservation into the training objective using a forward Euler scheme with a daily timestep. Building on this, \textit{April} further improved stability and performance by dynamically adjusting timesteps from daily to sub-daily intervals, effectively addressing numerical instabilities and ensuring compliance with the law of mass conservation. Tested across a range of lakes in the Midwestern USA, our models have demonstrated robust capabilities in predicting DO concentrations. \textit{Pril} and \textit{April} of utilizing physical processes are new frameworks that help us build models adhering to physical consistency and are not confined to the applications presented in this paper. We encourage more instantiations across various scientific and engineering disciplines to be proposed based on our work.

\section*{Acknowledgment}

This work was supported by the National Science Foundation (NSF) under grants 2239175, 2316305, 1942680, 1952085, 2021871, 2213549, 2126474, 2147195, 2425844, 2425845, and 2430978, the USGS awards  G21AC10564 and G22AC00266, and the NASA grant 80NSSC24K1061. Yiqun Xie gratefully acknowledges the support of Google’s AI for Social Good Impact Scholars program. This research was also supported in part by the University of Pittsburgh Center for Research Computing through the resources provided. 

% Yanhua Li was supported in part by NSF grants IIS-1942680 (CAREER), CNS-1952085 and DGE-2021871.
% Paul Hanson was supported by NSF grant 2213549
% Yiqun Xie is supported in part by the National Science Foundation under Grant No. 2126474, 2147195, 2425844 and 2430978, and Google's AI for Social Good Impact Scholars program.

% \clearpage

\bibliographystyle{IEEEtran}
\bibliography{mybibliography}

\end{document}